\documentclass[10pt,twocolumn,letterpaper]{article}

\usepackage{cvpr} 

\usepackage[utf8]{inputenc} 
\usepackage[T1]{fontenc}    
\usepackage{url}            
\usepackage{booktabs}       
\usepackage{amsmath,amssymb,amsfonts,bm,dsfont}      
\usepackage{nicefrac}       
\usepackage{microtype}      

\usepackage{algorithmic}
\usepackage{graphicx}
\usepackage{textcomp}

\usepackage{tikz}
\usepackage{pgfplots} 

\pgfmathdeclarefunction{gauss}{3}{%
	\pgfmathparse{1/(#3*sqrt(2*pi))*exp(-((#1-#2)^2)/(2*#3^2))}%
}

\usepgfplotslibrary{fillbetween}
\usetikzlibrary{patterns}
\pgfplotsset{compat=1.12} 

\usepackage{tcolorbox}

\usepackage{xcolor}

\definecolor{cvprblue}{rgb}{0.21,0.49,0.74}
\usepackage[pagebackref,breaklinks,colorlinks,citecolor=cvprblue]{hyperref}

\def\paperID{12834} 
\def\confName{CVPR}
\def\confYear{2024}

\title{May the Noise be with you:\\ Adversarial Training without Adversarial Examples}

\author{%
 Ayoub Arous$^{1}$, Andrés F López-Lopera$^{2}$,  \\
Nael Abu-Ghazaleh$^3$, Ihsen Alouani$^4$ \\ 
$^1$ Division of Engineering, New York University (NYU) Abu Dhabi, UAE \\ 
$^2$ CERAMATHS, UPHF, Valenciennes, France \\
$^3$ University of California Riverside, CA, USA
 \\
$^4$ CSIT, Queen’s University Belfast, UK\\
}

\begin{document}

\maketitle
\begin{abstract}

While the vulnerability of machine learning (ML) models to adversarial attacks has been thoroughly investigated by the community, it continues to be a persistent threat that undermines the trustworthiness of ML systems. Adversarial training (AT) remains the first line of defense used to harden ML models.  It works by solving a complex min-max optimization when training the network on adversarial examples. On the other hand, stochastic-based certifiable defenses like "randomized smoothing" introduce randomness into the input to reduce the attack surface and provide probabilistic guarantees of robustness. 
In this paper, we investigate the following question: \emph{Can we obtain adversarially-trained models without training on adversarial examples?} 
    
Our intuition is that training a model with inherent stochasticity, i.e., optimizing the parameters by minimizing a stochastic loss function, yields a robust expectation function that is non-stochastic. In contrast to related methods that introduce noise at the input level, our proposed approach incorporates inherent stochasticity by embedding Gaussian noise within the layers of the NN model at training time. 
We model the propagation of noise through the layers, introducing a closed-form stochastic loss function that encapsulates a noise variance parameter. Additionally, we contribute a formalized noise-aware gradient, enabling the optimization of model parameters while accounting for stochasticity. Our experimental results confirm that the expectation model of a stochastic architecture trained on benign distribution is adversarially robust. Interestingly, we find that the impact of the applied Gaussian noise's standard deviation on both robustness and baseline accuracy closely mirrors the impact of the noise magnitude employed in adversarial training.  Our work contributes adversarially trained networks using a completely different approach, with empirically similar robustness to adversarial training.  We hope that further exploration of this alternative may uncover to advantages in terms of both robustness and training time.

\end{abstract}    
\section{Introduction}

While ML models achieved unprecedented success across a diverse spectrum of applications, including critical domains, their vulnerability to adversarial attacks~\cite{pgd,CW, fgsm, tramer2020adaptive, Alouani2024} remains a significant concern. These attacks introduce bounded-magnitude perturbations into the model's input, that are maliciously tailored to force the output to a wrong label. Particularly, in safety-critical and security-sensitive contexts, these attacks represent a notable threat that undermines system security and safety and erodes the trustworthiness of ML.

Several defenses have been proposed against adversarial attacks, which can be classified into heuristic defenses and certified defenses~\citep{xue2020machine, sok_SP, Surv_AT}. Heuristic defenses focus on practical effectiveness. The widely used state-of-the-art heuristic defense is adversarial training (AT)~\citep{madry2019deep}. AT enhances the robustness of models intrinsically by exposing it to adversarial examples in the training data.  
Thus, adversarially trained models maintain integrity under adversarial attacks under a given noise magnitude. Mathematically, AT is formulated as a min-max problem, searching for the best solution to the worst-case optimum. Empirical results highlight the effectiveness of projected gradient descent (PGD) based adversarial training in achieving state-of-the-art accuracy against various $L_{\infty}$ attacks. 


Another approach is to provide complete or incomplete verification bounds towards provably robust models. Some of these techniques involve model certification through the formulation of an adversarial polytype and the establishment of its upper bound using convex relaxations \cite{alphabeta,betacrown}.   Robustness verification approaches have worst-case exponential time complexity due to the hardness of verification \cite{sok_certif}.
Within the body of certified robustness, we are particularly interested in probabilistic methods such as \textit{randomized smoothing} based approaches \cite{snP2019_certif,cohen2019certified,liu2017}. These approaches involve input or feature transformations, mainly through additive noise~\citep{li2019certified,strauss2017ensemble}, to mitigate adversarial effects in the data or feature space.  Liu et al. \cite{liu2017} suggest randomizing the entire DNN and predicting using an ensemble of multiple copies of the DNN. Lecuyer et al. \cite{snP2019_certif} also propose adding random noise to the first layer of the DNN and estimating the output using a Monte Carlo simulation. 
All these approaches inject stochastic noise to the input or the model \textbf{at inference time} and require \textit{computational expensive simulations} to infer the output which limits their practicality. Importantly, the main reason behind these limitations stems from the lack of analytical modeling for the noisy behavior of models.  

\begin{figure}[t!]
    \centering
    {\includegraphics[width=\columnwidth]{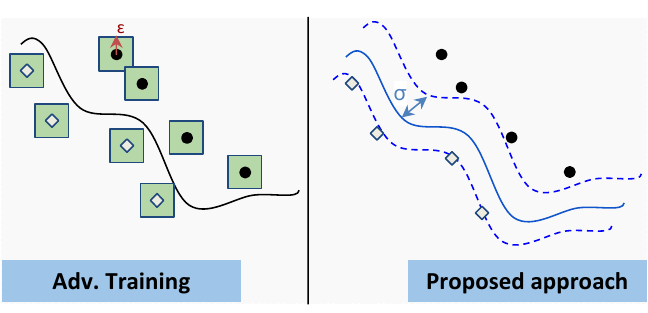}}
    \caption{Illustration of our intuition, in comparison to AT. While AT fits the model under an $\ell_p$-norm ball noise around the input samples to distance the decision boundary from the data distribution, we propose to optimize the model's parameters under stochastic behavior of the model itself. This will guarantee a distance from the expectation of the model and the data distribution. }
    \label{fig:intuition}
\end{figure}

\begin{figure*}[!h]
    \centering
    {\includegraphics[width=\textwidth]{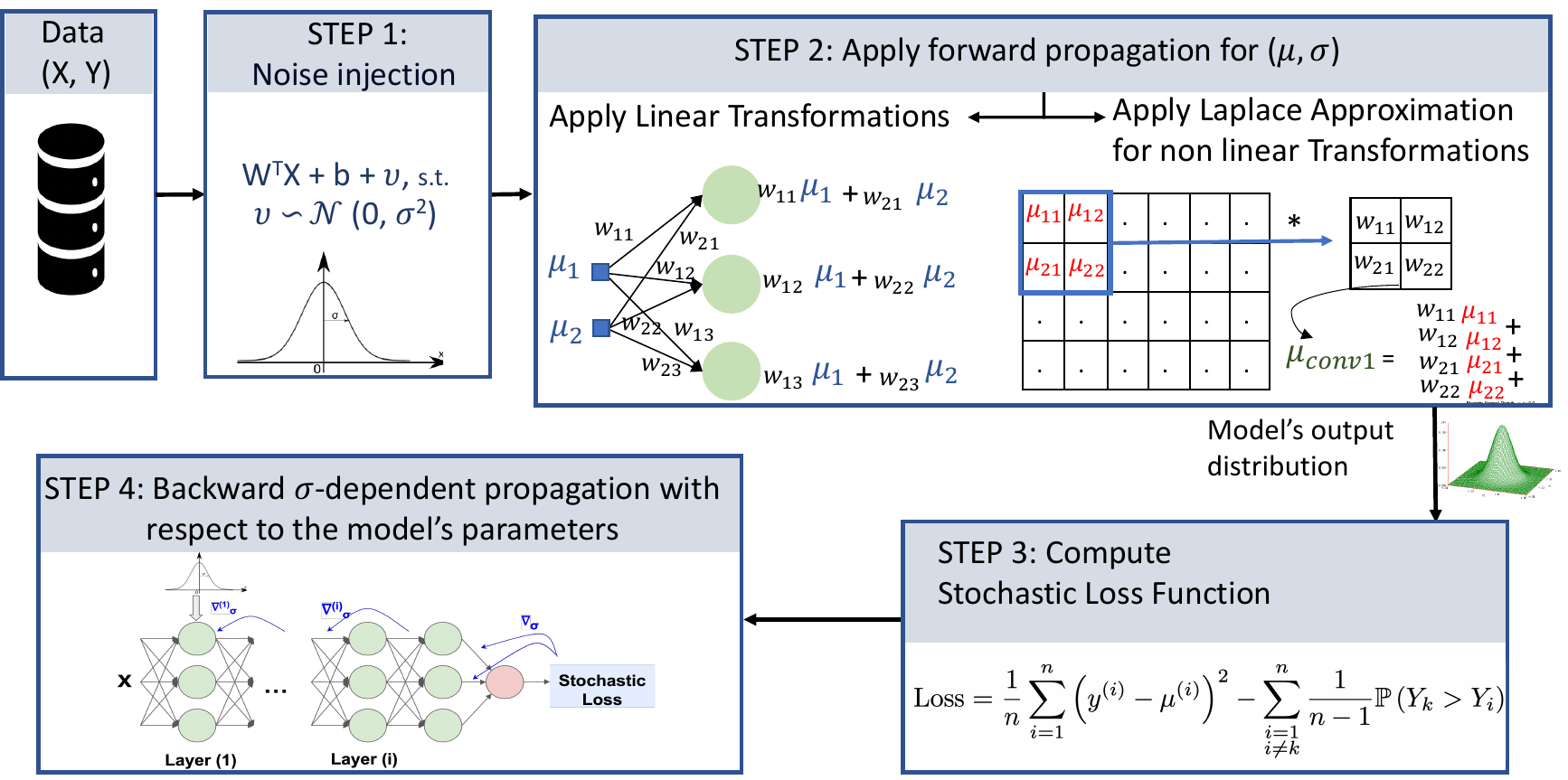}}
    \caption{ An overview of our proposed stochastic training methodology. }
    \label{fig:overview}
\end{figure*}

Motivated by bridging the gap between the two defense categories (i.e., AT and randomized techniques), we propose a new approach which trains adversarially robust models by leveraging random noise \textbf{at training time}. Our intuition is that training a model with inherent stochasticity, i.e., optimizing the parameters by minimizing a stochastic loss function, yields a robust expectation function (non-stochastic model). As illustrated in Figure \ref{fig:intuition}, we are inspired by AT. However, instead of using the noise as an $\ell_p$-norm ball in the neighborhood of the data samples, we add the noise to the model itself during the training. We propose that training a stochastic model while taking noise into account converges to an expectation function that maximizes the decision boundary distance from the data samples. To do so without expensive Monte-Carlo simulations, we need a closed form of the Loss function, which takes into account the noise distribution. Therefore, we consider a zero-centered Gaussian noise in a layer pre-activation and analytically model its propagation through the NN. Through this propagation, we obtain a closed-form expression of a stochastic loss function, which is parameterized by the standard deviation of the initial noise. Additionally, we express a stochastic Jacobian distribution and back-propagate the noise-aware gradient to update the model parameters. At inference, the model is then inferred as the expectation of the stochastically-trained model. The proposed noise-aware training methodology is illustrated by Figure~\ref{fig:overview}.

 Our experiments show that the expectation of the trained stochastic model is an adversarially-robust model. Interestingly, we observe the same trend in robustness as we vary the standard deviation ($\sigma$) of the training stochasticity. This observation establishes that a higher noise distribution results in higher robustness against adversarial attacks, which draws a parallel with the impact of the training noise magnitude in AT.  We believe that this represents a new methodology for adversarial hardening with attractive properties.

\noindent\textbf{Contributions.} The contributions of this work are summarized as follows: 
\begin{itemize}
    \item We propose a new randomization-based technique to train adversarially robust models without min-max optimisation, i.e., without training on adversarial examples. At training time, we consider additive Gaussian noise injected to the first layer of the model with 0-mean and a standard deviation $\sigma$. 
    
    \item To train the model under noise, we propose a closed-form expression of the loss as a function of the propagated noise. We address this problem by handling the non-linearity within the NN layers using Laplace approximation. By propagating the distribution through the layers, we finally integrate it in the expression of our loss function. This allows us to model the output distribution \textbf{without the need of Monte Carlo simulation}.  The hardened model achieves similar robustness to AT.

    \item  We express a closed form of the gradient as a function of the noise distribution. Interestingly, this formulation allows the optimization of the noise parameter during training. Therefore, the second contribution of the paper relies on the consideration of the variance as a learnable parameter, rather than a hyper-parameter as it is often assumed in the state-of-the-art.

    \item We show that using the closed form and gradient of the stochastic model, we can be leveraged to build adaptive attacks against defenses that rely on random noise at inference time. 
\end{itemize}


\newcommand{\ReLU}{\operatorname{ReLU}}
\definecolor{mygreen}{rgb}{0, 0.5, 0}

\section{Proposed Approach}

Towards enhancing the robustness of ML models, we propose a novel approach that bridges the gap between adversarial training and stochastic-based defenses. Through the min-max optimization, AT  in essence trains the model to fit a decision boundary that is as far as possible from the data distribution, while still providing the correct classification. Aiming for the same objective, our approach is illustrated in Figure \ref{fig:intuition} and is composed of two main parts: \\
\noindent\textbf{(i) At training time:} we consider a model that has a stochastic component by injecting a Gaussian noise pre-activation of the first layer. We propagate the distribution through the model to have a loss function that is parametrized by the noise distribution's parameters (essentially the standard deviation $\sigma$). Minimizing a loss under a \textbf{stochastic decision boundary} can be seen as transposing the AT problem from the data to the model. In fact, training a model with inherent stochasticity means finding the parameters that distances the decision boundary from the training samples. \\
\noindent \textbf{(ii) At inference time:} Once the stochastic model converges, the inference is performed with the expectation of the trained model, i.e., withdrawing the Gaussian noise and keeping the model's parameters. This (deterministic) model has adversarial robustness that is dependant on the noise parameter $\sigma$, which is analogous to the relation between robustness in AT and the training noise magnitude.


The training problem for a stochastic model can then be expressed as follows: 

\begin{equation}
    \min _{W} \mathbb{E}_{(x, y) \sim \mathcal{D}}\left[ \mathcal{L}_{\sigma}( W, f(x), y)\right]
\end{equation}

Where $\mathcal{L}_{\sigma}$ is the stochastic loss function, $W$ indicates the parameters of the classifier $f(\cdot)$,   which is parametrized by the noise standard deviation $\sigma$, and $(x, y) \sim \mathcal{D}$ represents the training data sampled from a
distribution $\mathcal{D}$.

The core part that enables our approach lies in modeling the propagation of noise through these layers, culminating in a closed-form stochastic loss function that encapsulates a noise variance parameter. This departure from the norm not only introduces a layer of inherent stochasticity but also equips the model with the ability to adapt to perturbations in the input space. Additionally, we present a formalized stochastic gradient, enabling the optimization of model parameters while adeptly accounting for the model's stochasticity.

The proposed approach not only enhances the model's robustness against adversarial attacks but also avoids the inference computational bottleneck associated with traditional stochastic-based defenses. 

Propagating uncertainties through an NN can be challenging, particularly when dealing with the non-linear layers of the network. The difficulty arises from the fact that the distribution of the network output depends not only on the input distribution but also on the weights of the network, which are typically unknown and need to be learned from data. Moreover, this problem is challenging due to the lack of explicit output modeling that takes the stochastic aspect into account. 


\subsection{Forward propagation}
\label{sec:modelingNoise:subsec:forwardProp}
To define the stochastic loss which includes explicitly within its expression the injected noise parameter, we need to propagate this injected noise which follows a Gaussian distribution, through the NN. To do so, we need to understand the transformations within our NN and handle the non-linear ones. Here, we consider convolution neural networks (CNNs), which include affine and linear transformations (convolution and fully connected layers) and a non-linear transformation for the activation function. For the former transformations, due to the properties of Gaussian distributions, we can show that the distribution of the outcome after applying them remains Gaussian~\citep{Rasmussen2005GP}. In contrast, for the non-linear case, for instance, assuming the ReLU transformation, the outcome is not Gaussian anymore. However, it is still possible to promote Gaussianity by considering a Laplace approximation. Therefore, since Gaussianity is preserved in each layer of the NN model, the output will also be Gaussian where the variance will depend on the noise variance.

In the following, for simplicity, we will consider that $\bm{X} \in \mathbb{R}^{d_1 \times d_2}$ (e.g. a one-channel 2D input image) but intuitions can be generalized to 3D tensor objects with a more cumbersome notation.

\paragraph{Convolution layer.}

The convolutional transformation $f_{\bm{W}, \bm{b}}(\cdot)$, with weight and bias parameters $\bm{W}$ and $\bm{b}$ (respectively), is given by:
\begin{equation}
    f_{\bm{W}, \bm{B}}(\bm{X})
    = \bm{X} \ast \bm{W} + \bm{B},
    \label{eq:2Dconvolution}
\end{equation}
where $\ast$ is the convolution operator. Here, $\bm{W}$ is an $m_{1} \times m_{2}$ matrix corresponding to the convolutional filter. The size of the matrix $\bm{B}$ will depend on the 2D convolution. Here, we assume that $\bm{B}$ is an $p_1 \times p_2$ matrix.


The 2D convolution in~\eqref{eq:2Dconvolution} can be generally written in the following matrix form:
\begin{equation*}\label{eq:2dconv}
    \bm{y}_{\text{conv}} = \bm{A}^\top \bm{x} + \bm{b},
\end{equation*}
where $\bm{x} := \operatorname{vect}(\bm{X}) \in \mathbb{R}^{d_1 \cdot d_2}$ and $\bm{b} := \operatorname{vect}(\bm{B}) \in \mathbb{R}^{p_1 \cdot p_2}$. The matrix $\bm{A} \in \mathbb{R}^{(d_1 \cdot d_2) \times (p_1 \cdot p_2)}$ is constructed from $\bm{W}$ taking into account the multiplications involved in the 2D convolution. 

We seek to inject an additive Gaussian perturbation $\bm{\nu} \sim \mathcal{N}(\bm{0}, \sigma^2 \bm{I})$ at the first layer of the convolutional architecture, i.e. $\bm{Z} = \bm{y}_{\text{conv}} + \bm{\nu}$. Because of the linearity, we can show that $\bm{Z}$ is also Gaussian-distributed:
\begin{equation}
    \bm{Z} \sim \mathcal{N}(\bm{\mu}_z, \sigma^2 \bm{I}).
    \label{eq:GaussianDist}
\end{equation}
with $\bm{\mu}_z = \bm{A}^\top \bm{x} + \bm{b}$. We observe that $\bm{Z}$ has elements $Z_{i} \sim \mathcal{N}(\mu_i, \sigma^2)$, for $i = 1, \ldots, p$ with $p = p_1\cdot p_2$, where $\mu_i = \bm{a}_i^\top \bm{x} + b_i$ with $\bm{a}_i$ the $i$-th column of the matrix $\bm{A}$.

We must remark that a fully connected layer is also expressed by an affine transformation. Therefore, the aforementioned formulation can also be used for establishing the Gaussian distribution of the outcome.

%


\paragraph{ReLU Transformation.}
Now, our interest is to determine the distribution of the outcome after applying the ReLU transformation, i.e. the distribution of $Y_{i} = \operatorname{ReLU}(Z_i) := \max(0, Z_i)$, for $i = 1, \ldots, p$. To simplify our analysis, we drop the index $i$ in the following, i.e. we consider $Y = \operatorname{ReLU}(Z)$ with $Z \sim \mathcal{N}(\mu, \sigma^2)$.

Due to the non-linearity, $y$ is not Gaussian but truncated Gaussian distributed with probability density function (pdf) given by \cite{Botev2019}
%
\begin{align}
    f_{Y}\left(\xi ; \mu, \sigma\right)
    =& \frac{1}{\sigma \eta} \varphi\left( \frac{\xi - \mu}{\sigma}\right) \mathds{1}_{\xi > 0},
    \label{eq:trGaussianPDF}
\end{align}
where $\varphi\left( \xi\right) = \frac{1}{\sqrt{2\pi}} \exp \left(-\frac{\xi^2}{2} \right)$ is the pdf of the standard normal distribution, and $\eta = 1 - \Phi\left(-\frac{\mu}{\sigma} \right)$ with $\Phi(\xi) = \frac{1}{2} [1 + \operatorname{erf}(\xi/\sqrt{2}) ]$ the cumulative distribution function (cdf) and $\operatorname{erf}(\cdot)$ the Gaussian error function. The indicator function $\mathds{1}_{\xi > 0}$ is equal to one if $\xi > 0$ and zero otherwise.\\ 
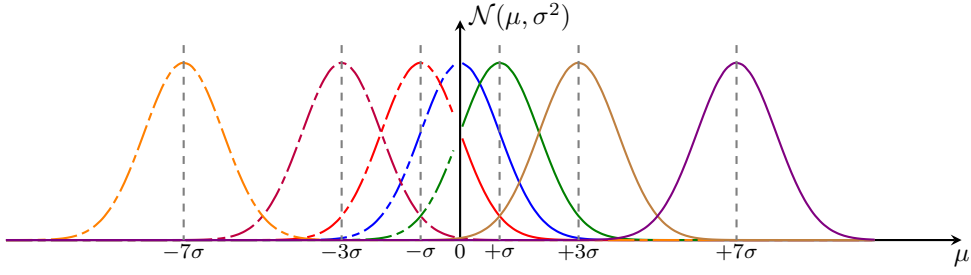
\begin{figure*}[t!]
	\centering
	\begin{tikzpicture}
		\def\S{1};
            \def\m{0}
		\def\h{gauss(\m,\m,\S)};
		
            \def\xmin{-11.5}
		\def\xmax{10.5};
		\def\ymin{0};
		\def\ymax{0.45};
  
		\begin{axis}[scale = 0.8,
			every axis plot post/.append style={
				mark=none,
                domain={\xmin}:{\xmax}, samples=100, smooth},
			xmin=\xmin, xmax=\xmax,
			ymin=\ymin, ymax=\ymax,
			axis lines=middle,
			axis line style=thick,
			enlargelimits=upper, 
			ticks=none,
			xlabel=$\mu$,
			every axis x label/.style={at={(current axis.right of origin)},anchor=north},
			width=\textwidth, height=0.3*\textwidth,
			clip=false
			]
			\filldraw (0, \ymax) node[above right] {$\mathcal{N}(\mu, \sigma^2)$};

                \addplot[blue, thick, name path=B, restrict x to domain=-0.1:\xmax] {gauss(x, \m, \S)};
                \addplot[blue, thick, dash pattern={on 10pt off 2pt on 3pt off 2pt}, restrict x to domain=\xmin:0] {gauss(x, \m, \S)};
                \addplot[red, thick, name path=B, restrict x to domain=-0.1:\xmax] {gauss(x, -\S, \S)};
                \addplot[red, thick, dash pattern={on 10pt off 2pt on 3pt off 2pt}, restrict x to domain=\xmin:0] {gauss(x, -\S, \S)};
                \addplot[mygreen, thick, name path=B, restrict x to domain=-0.1:\xmax] {gauss(x, \S, \S)};
                \addplot[mygreen, thick, dash pattern={on 10pt off 2pt on 3pt off 2pt}, restrict x to domain=\xmin:0] {gauss(x, \S, \S)};
			\addplot[purple, dash pattern={on 10pt off 2pt on 3pt off 2pt}, thick, name path=C] {gauss(x, -3*\S, \S)};
                \addplot[orange, dash pattern={on 10pt off 2pt on 3pt off 2pt}, thick, name path=D] {gauss(x, -7*\S, \S)};
			\addplot[brown, thick, name path=E] {gauss(x, 3*\S, \S)};
                \addplot[violet, thick, name path=F] {gauss(x, 7*\S, \S)};

			\addplot[dashed,thick,gray]
			coordinates {(\m,0) (\m,0)}
			node[below=-2pt, black] {\footnotesize $0$};
                \addplot[dashed,thick,gray]
			coordinates {(-\S,1.1*\h) (-\S,0)}
			node[below=-2pt, black] {\footnotesize $-\sigma$};
                \addplot[dashed,thick,gray]
			coordinates {(\S,1.1*\h) (\S,0)}
			node[below=-2pt, black] {\footnotesize $+\sigma$};
                \addplot[dashed,thick,gray]
			coordinates {(3*\S,1.1*\h) (3*\S,0)}
			node[below=-2pt, black] {\footnotesize $+3\sigma$};
                \addplot[dashed,thick,gray]
			coordinates {(-3*\S,1.1*\h) (-3*\S,0)}
			node[below=-2pt, black] {\footnotesize $-3\sigma$};
			\addplot[dashed,thick,gray]
			coordinates {(7*\S,1.1*\h) (7*\S,0)}
			node[below=-2pt, black] {\footnotesize $+7\sigma$};
                \addplot[dashed,thick,gray]
			coordinates {(-7*\S,1.1*\h) (-7*\S,0)}
			node[below=-2pt, black] {\footnotesize $-7\sigma$};
			
		\end{axis}
	\end{tikzpicture}
	\caption{Censoring effect of the ReLU function (dashed red line) on the pdf of $Y \sim \mathcal{N}(\mu, \sigma^2)$. Curves are displayed for $\sigma^2 = 1$ and for different values of $\mu$. The censored parts of the pdfs are displayed in dash-dotted lines.}
	\label{fig:censorEffectReLU}
\end{figure*}
From the normalizing constant $\eta$, we can observe that:
\begin{itemize}
    \item $\eta \xrightarrow[\mu \to + \infty]{} 1$, implying that $f_{y}\left(\xi ; \mu, \sigma\right)$ tends to an untruncated Gaussian distribution. 
    \item $\eta \xrightarrow[\mu \to - \infty]{} 0$, implying that $f_{y}\left(\xi ; \mu, \sigma\right) \to \delta(0)$ (as a result of censoring the negative values) \cite{Beauchamp2018rectifiedGaussianVariables}, which can be approximated by an untruncated Gaussian distribution with $\mu = 0$ and $\sigma^2 \to 0$.
\end{itemize}
Numerically, these cases are exhibited when $\mu > 3\sigma$ or $\mu < - 3\sigma$, respectively. For the case when $|\mu| \leq 3\sigma$, we consider a Laplace approximation to promote Gaussianity. Note that the interval $[-3\sigma, +3\sigma]$ is also justified by the 99.7\% coverage of the Gaussian distribution. Figure~\ref{fig:censorEffectReLU} illustrates the censoring effect of the ReLU function.



To be able to apply a Laplace approximation on the pdf in~\eqref{eq:trGaussianPDF}, we need to focus only on the positive region. This will ensure continuity when considering the Taylor expansion. Suppose $h(\xi) = \ln(f_Y(\xi ; \mu, \sigma))$. The second-order Taylor expansion around the maximum of $h(\xi_0)$ is given by
\begin{equation*}
    h(\xi) \approx
    h(\xi_0)
    + h''(\xi_0) \frac{(\xi-\xi_0)^{2}}{2},
\end{equation*}
where $h''(\xi) = - \sigma^{-2}$.
Then,
\begin{align}
    f_Y(\xi ; \mu, \sigma^{2})
    \approx \exp(h(\xi))
    \propto \exp\left(-\frac{(\xi-\xi_0)^{2}}{2 \sigma^2}\right).
    \label{eq:LaplaceApprox}
\end{align}
For the mean $\xi_0$, we consider the mode of the pdf in~\eqref{eq:trGaussianPDF} (see Figure~\ref{fig:censorEffectReLU}
for an illustration):
\begin{equation*}
    \xi_0 
    = \underset{\xi}{\arg\max} f_Y(\xi ; \mu, \sigma^{2})
    =
    \begin{cases}
    0, & \text{if } -3\sigma \leq \mu < 0, \\
    \mu, & \text{if } 0 \leq \mu \leq 3\sigma.
    \end{cases}
\end{equation*}
Observe that~\eqref{eq:LaplaceApprox} has a closed-form which depends on the noise variance $\sigma^2$ of the additive Gaussian perturbation $\bm{z}$. 

By denoting $\bm{y}_{\operatorname{ReLU}}$ the vector composed by the outcomes after applying the ReLU transformation, we can establish the distribution: 
\begin{equation}
    \bm{y}_{\operatorname{ReLU}} \sim \mathcal{N}(\bm{\mu}_{\text{ReLU}}, \operatorname{diag}(\sigma_1^2, \ldots, \sigma_{p}^2)),
    \label{eq:GaussianDistReLU}
\end{equation}
where $\bm{\mu}_{\text{ReLU}}$ is the mean vector with elements  $\mu_{\text{ReLU},i} = \mu_i$ if $\mu_i \geq 0$, and zero otherwise. The variances $\sigma_i^2$ are equal to $\sigma^2$ (noise variance) if $\mu_i \geq -3\sigma$, and close to zero otherwise.

For the next convolutional layer, we can follow the same procedure but considering the distribution in~\eqref{eq:GaussianDistReLU} rather than the one in~\eqref{eq:GaussianDist}. As a result, we will be able to propagate noise sequentially across the NN. We must remark that, while the weights of the NN model interplay a key role in the definition of the mean of the output distribution (see interaction in~\eqref{eq:GaussianDist}), the variances will only depend on the noise variance $\sigma^2$.


%

\paragraph{Maxpooling transformation.}
Pooling layers are used to reduce the dimensions of the feature maps. Thus, it reduces the number of parameters to learn and the amount of computation performed in the network. The pooling layer summarizes the features present in a region of the feature map generated by a convolution layer. To do so, we conserve features of the pooling layer by taking the maximum of the mean value in each region which we call the mean-pooling layer.


\subsection{Stochastic loss function}
\label{subsec:stochLossFun}
The injection of Gaussian noise in the NN motivates the construction of a loss function that can take into account the stochasticity in the model. For this reason, we seek to provide a loss function that jointly enhances the model's accuracy in the training phase and increases the score of predicting the true label. 

Let $\bm{Y} \sim \mathcal{N}((\mu_1, \ldots, \mu_n)), \operatorname{diag}(\sigma_1^2, \ldots, \sigma_n^2))$ be the Gaussian distribution of the NN output, with $n \in \mathbb{N}$ corresponding to the number of neurons at the latest layer (i.e. the number of classes). Assume that $k$ is the true label for the specific task prediction. Then, we define the stochastic loss function as:
\begin{equation*}
    \operatorname{Loss} = \frac{1}{n} \sum_{i=1}^{n} \left(y^{(i)}-{\mu}_{i} \right)^2 - \sum_{\substack{i = 1 \\ i \neq k}}^n \frac{1}{n-1} \mathbb{P}\left(Y_{k} > Y_{i}\right),
\end{equation*}
where, for $i = 1, \ldots, n$, $y^{(i)}$ and ${\mu}^{(i)}$ are the ground truth label and the mean of the $i$-th neuron. 

Note that the loss is decomposed into two parts. The first term, which corresponds to the mean squared error (MSE), seeks to improve the accuracy of the model (i.e. the accuracy of the mean as a predictor). On the other hand, the second term seeks to maximize the probability of predicting the true label, which will also enhance the accuracy of the model. 

For the case of multivariate Gaussians, i.e. $\begin{bmatrix} X_1, X_2 \end{bmatrix}^\top \sim \mathcal{N}\left(\bm{\mu}, \bm{\Sigma}\right)$ with mean vector $\bm{\mu} = \begin{bmatrix}\mu_1, \ \mu_2 \end{bmatrix}^\top$ and covariance matrix $(\bm{\Sigma})_{1 \leq i,j \leq 2} = \sigma_{1,2}^2$, we have that
\begin{align*}
    \mathbb{P}(X_1 > X_2)
    &= \operatorname{CDF}_{X_2-X_1}(0)
    \\
    &= \frac{1}{2}\Bigg[1+\operatorname{erf}\Bigg(\frac{\mu_{1}-\mu_{2}}{\sqrt{2\left(\sigma_{1,1}^{2}+\sigma_{2,2}^{2}-2 \sigma_{1,2}\right)}}\Bigg)\Bigg].
\end{align*}
In our case, we need to consider the case where $X_1$ and $X_2$ are independent (see~\eqref{eq:GaussianDistReLU}).    
Therefore, using the aforementioned property for computing the probabilities $\mathbb{P}\left(Y_{i} > Y_{k}\right)$, and taking into account the independence between $Y_1, \ldots, Y_n$, the stochastic loss can be written as:
\begin{align*}
    \operatorname{Loss}
    = \operatorname{MSE}
    - \sum_{\substack{i = 1 \\ i \neq k}}^n \frac{1}{2n-2}\left[1+\operatorname{erf}\left(\frac{\mu_{i}-\mu_{k}}{\sqrt{2\left(\sigma_{i}^{2}+\sigma_{k}^{2}\right)}}\right)\right],
\end{align*}
with $\operatorname{MSE} = \frac{1}{n} \sum_{i=1}^{n} (y^{(i)}-{\mu}_i)^{2}$. We should note that our objective function involves the parameters of the Gaussian vector $\bm{Y}$ which have been previously computed in the forward propagation of the noise. As discussed in Section~\ref{sec:modelingNoise:subsec:forwardProp}, the means $\mu_1, \ldots, \mu_n$ will depend on the weights and biases of the NN, and the variances $\sigma_1^2, \ldots, \sigma_n^2$ will depend only on the noise variance $\sigma^2$. Therefore, the stochastic loss can be written as a function of $\bm{W}$, $\bm{B}$ and $\sigma$, i.e. $\operatorname{Loss}(\bm{W}, \bm{B}, \sigma)$, which allows establishing the optimization problem:
\begin{equation}
    (\bm{W}^\star, \bm{B}^\star) 
    = \underset{\bm{W}, \bm{B}}{\arg\min} \operatorname{Loss}(\bm{W}, \bm{B}, \sigma).
    \label{eq:optProb}
\end{equation}

\noindent \textbf{Remark.} It is worth mentionning that the expressed loss function also allows to define the follwing optimisation problem:
\begin{equation}
    (\bm{W}^\star, \bm{B}^\star, \sigma^{\star}) 
    = \underset{\bm{W}, \bm{B}, \sigma}{\arg\min} \operatorname{Loss}(\bm{W}, \bm{B}, \sigma).
    \label{eq:optProb2}
\end{equation}
Interestingly, this formulation make the injected noise variance $\sigma^2$ a learnable parameter that can be jointly optimised with the NN parameters. The optimization problem in~\eqref{eq:optProb2} can be seen as an increase in the dimensionality of the (deterministic) NN model by adding a new dimension to the hyperspace. Further details and experiments about this perspective can be found in Section \ref{sec:xyz} of the supplementary material.


\subsection{Backward propagation of the noise-aware gradient}
The next step is to find the partial derivatives of the stochastic loss function to enable the training by gradient back-propagation.  To do so, need to find the stochastic Jacobian to find the noise-aware updates of the parameters. More precisely we are going to use the chain rule to compute the $\sigma$-parametrised gradient. 

\paragraph{Backpropagation through the last layer} Assuming that we have a last fully connected layer with n neurons, weights $W$, inputs noise $\sigma_{out-1}$ and output noise $\sigma_{out}$ , we have the expression of the loss after propagating the noise distribution, which is a function of $\sigma_{out}$. Computing its derivative gives us the final expression of $\frac{\partial \operatorname{Loss}}{\partial W}$ which will be if we have $k^{th}$ neuron is the true label:

\begin{align*} 
    \frac{\partial \operatorname{Loss}}{\partial\sigma_{W_{i,j}}} 
    &= \frac{\partial \operatorname{Loss}}{\partial\sigma_{out_i}}\frac{\partial \operatorname{\sigma{out_i}}}{\partial\sigma_{W{i,j}}}
    \\
    &= \frac{\partial \operatorname{\sum_{\substack{i = 1 \\ i \neq k}}^n \frac{1}{2n-2}\left[1+\operatorname{erf}\left(\frac{\mu_{i}-\mu_{k}}{\sqrt{2\left(\sigma_{i}^{2}+\sigma_{k}^{2}\right)}}\right)\right]}}{\partial \sigma_{out_i}}  \frac{\partial \operatorname{\sigma{out_i}}}{\partial\sigma_{W{i,j}}}
    \\
    &=\frac{\sigma_{(output-1)_j}(\mu_k-\mu_i)\sigma_i}{(\sigma_{out_i}^2+\sigma_{out_k}^2)^{\frac{3}{2}}\sqrt{\Pi}} \exp{-\left( \frac{\mu_i-\mu_k}{2(\sigma_{out_i}^2-\sigma_{out_k}^2)}\right)}
\end{align*}

It is worth noticing that the gradient over the parameters is dependent on the standard deviation of the layer itself, which is the result of propagating the initial noise. Once we have the gradient over the last layer's parameters, the remaining process is a standard backpropagation via chain rule.


    \paragraph{Backpropagation through convolution layers.}
For the convolution layer, the calculation of the gradient of the output of the convolution with respect to its input is calculated as follows:\\
Let consider the $l^{th}$ layer of the convolution, $x$ with dimension $H*W$, a filter $w$ with dimensions $k_1*k_2$, $b^l$ is the bias, $f$ is the activation function of the $l^{th}$ layer. 
Denote $o_{i, j}^l = f\left(x_{i, j}^l\right)$ and $\delta_{i, j}^l = \frac{\partial \operatorname{Loss}}{\partial x_{i, j}^l}$. The backpropagation equations are as follows:
$$
\begin{aligned}
& x_{i, j}^l=\sum_m \sum_n w_{m, n}^l o_{i+m, j+n}^{l-1}+b_{i, j}^l \\
& \frac{\partial \operatorname{Loss}}{\partial x_{i^{\prime}, j^{\prime}}^l}=\sum_{m=0}^{k_1-1} \sum_{n=0}^{k_2-1} \delta_{i^{\prime}-m, j^{\prime}-n}^{l+1} w_{m, n}^{l+1} f^{\prime}\left(x_{i^{\prime}, j^{\prime}}^l\right) \\
& \frac{\partial \operatorname{Loss}}{\partial w_{m^{\prime}, n^{\prime}}^l}=\sum_{i=0}^{H-k_1} \sum_{j=0}^{W-k_2} \delta_{i, j}^l o_{i+m^{\prime}, j+n^{\prime}}^{l-1}
\end{aligned}
$$

For the fully connected layer, this is also straightforward due to the linearity of the operation.

\paragraph{Backpropagation of the maxpooling and activation function.}
For the pooling layer, the process is the same as the conventional back-propagation the gradient. The gradient is only considered at the maximum of each region and then we are performing a padding task at each point to reshape the matrices and return to the same dimension at the input of this layer. The backpropagation through ReLU issimilar to conventional models.

\section{Empirical Evaluation}

\subsection{Setup}
We conducted experiments on both MNIST and CIFAR-10 datasets to empirically evaluate the performance of our methods. For MNIST dataset, we train a Lenet-5 (3 Convolution layers and 2 fully connected layers), with ReLU activation functions and max-pooling layers. Additionally, there were two fully connected layers, each with a size of 200. For CIFAR-10, we trained a CNN with five convolutional layers, each followed by ReLU and MaxPooling layers, and three fully connected layers. 

We proceeded with classification and accuracy measurement across various fixed standard deviation values, considering both forward and backward passes while accounting for the model's stochasticity within the weights optimization. 

We evaluate the model's robustness using Projected Gradient Descent (PGD) attack \cite{pgd}, as a state-of-the-art attack. Other results using FGSM can be found in the supplementary material.

\subsection{Results}
\begin{figure}[htp]
\centering
\includegraphics[width=\linewidth]{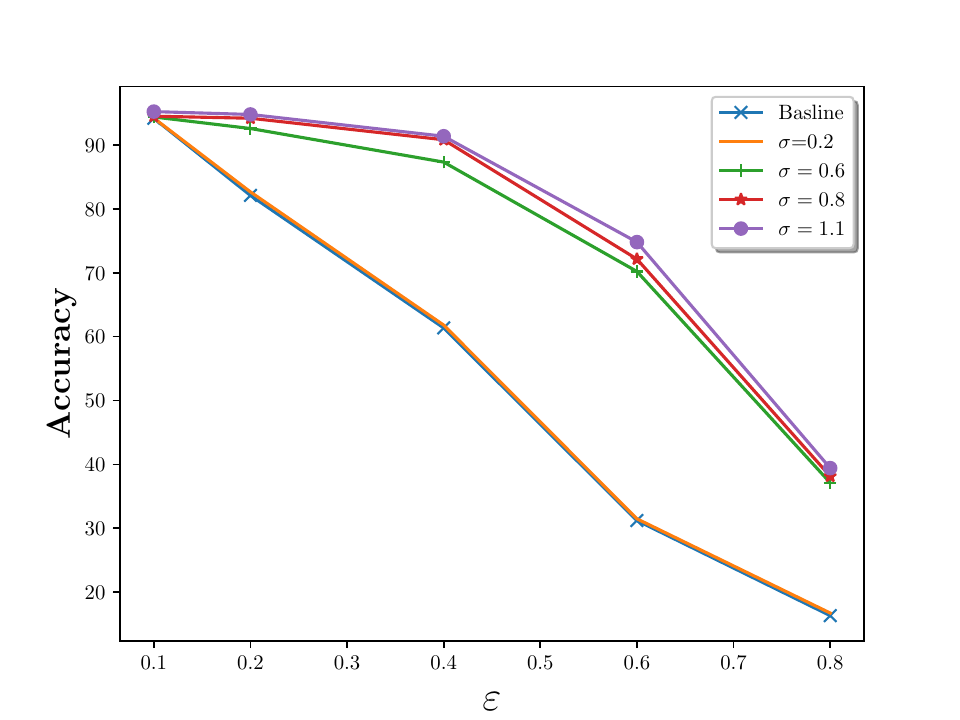}
\caption{Adversarial robustness of the stochastically trained model comparatively with the baseline model for MNIST under PGD attack.}
\label{Fig_pgdmnist}
\end{figure}

\begin{figure}[htp]
\centering
\includegraphics[width=\linewidth]{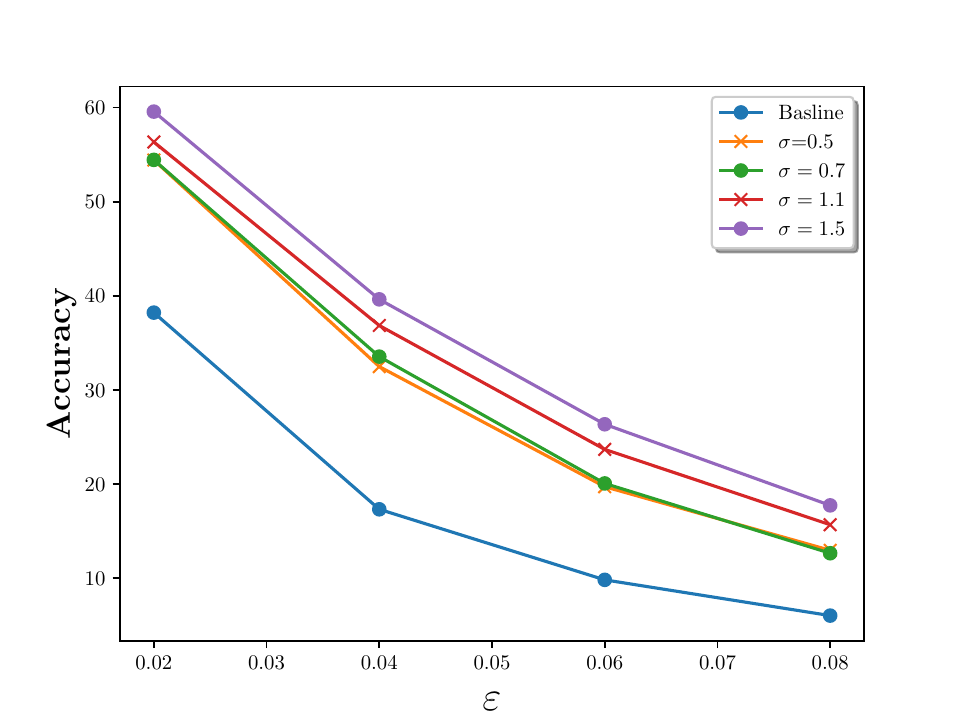}
\caption{Adversarial robustness of the stochastically trained model comparatively with the baseline model for CIFAR10 under PGD attack.}
\label{Fig_pgdcifar}
\end{figure}

\noindent\textbf{Impact on robustness.} Figure \ref{Fig_pgdmnist} shows the adversarial robustness of the stochastically trained model with different $\sigma$ levels comparatively with the baseline model for MNIST under PGD attack. \ref{Fig_pgdmnist} shows that higher $\sigma$ results in more robustness to adversarial noise.  The same trend has been obseved in Figure \ref{Fig_pgdcifar} which depicts the adversarial robustness of the stochastically trained model comparatively with the baseline model for CIFAR10 under PGD attack.





\begin{figure}[htp]
\centering
\includegraphics[width=\linewidth]{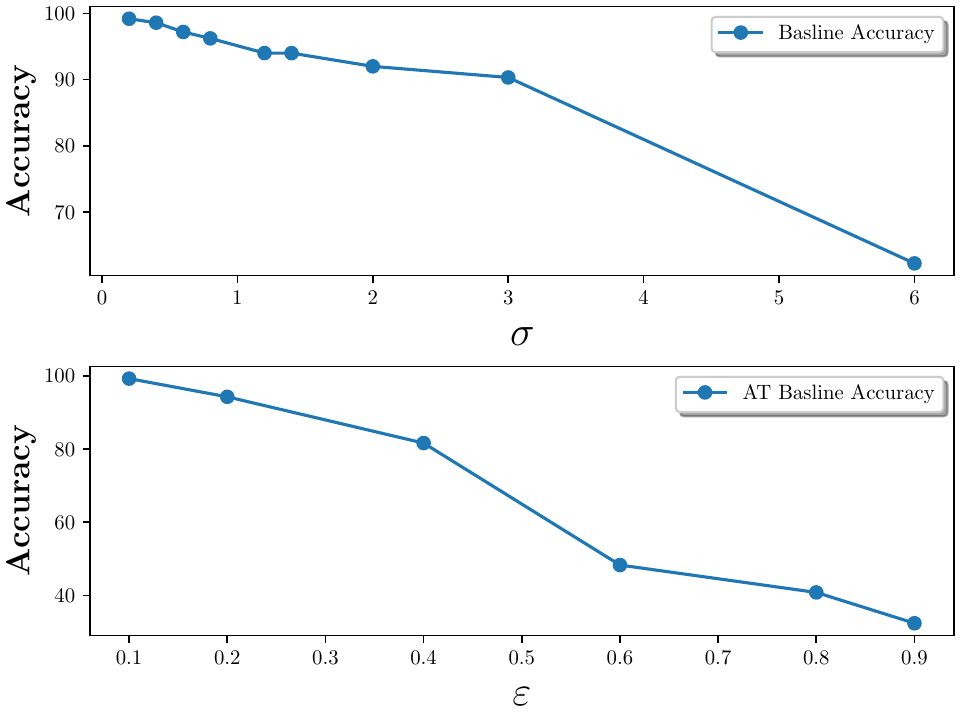}
\caption{Baseline accuracy of models trained with our approach while varying the standard deviation of the noise (top); and adversarially trained model while varying the adversarial noise budget used for AT (bottom) for MNIST.} 
\label{fig:acc_sigma}
\end{figure}

\noindent \textbf{Impact on baseline accuracy-- } In this experiment we wanted to investigate the impact of the noise magnitude at training time on the baseline accuracy of the model. We compare these results with the (well known) impact of adversarial noise budget on AT on the model's accuracy. 
The results are shown in Figure \ref{fig:acc_sigma}, which illustrates a decline in baseline accuracy as the noise level increases, a trend consistently observed during adversarial training accordingly for adversarial noise. While the objective of this experiment is not to quantitatively compare AT and stochastic training, it draws an interesting parallel which confirms the analogy we illustrated in Figure \ref{fig:intuition}.

\section{Adaptive attacks against Inference Time Randomization Techniques}
\label{sec:AdaptiveAttacks}

In this section, we investigate if the stochasticity-aware loss function and gradient can be used to build  adaptive attacks against defense strategies that use randomness at inference time as a defense. Given a randomized model $\widetilde{f}_{\sigma, W}(\cdot)$, which injects random noise to the first layer such as PixelDP \cite{Bai2019NIPS} at inference time, our objective is to generate adversarial noise under a white-box setting; The attacker is assumed to have total access to the model's architecture, the parameters as well as to the defender's noise.  We assume the attacker has access to the closed form of the stochastic loss and its gradient. The attacker implements a backpropagation of the gradient. The problem is therefore formalised as follows: 
\begin{equation}
    \underset{ \delta < \varepsilon }{\max} \operatorname{Loss}(x+\delta, y, \widetilde{f}, W, \sigma).
\end{equation}

The loss function contains the noise parameter $\sigma$, and therefore, the adversarial example can be generated as follows: 
\begin{equation}\label{s-fgsm}
  x^{adv} = x+\varepsilon \operatorname{sign}\left(\nabla_x Loss(x, y, \widetilde{f}, W, \sigma)\right) .
\end{equation}

To back-propagate the gradient to the input to implement the method described in Equation \ref{s-fgsm}, the noise-aware gradient of the loss with respect to the input is expressed as follows:

\begin{align*}
    \frac{d \operatorname{Loss}}{d\bm{x}}
    = \sum_{i=1}^n \frac{d \operatorname{Loss}}{d\mu_i} \frac{d\mu_i}{d\bm{x}}
\end{align*}
where for all $i = 1, \ldots n$,
\begin{equation*}
    \frac{d \operatorname{Loss}}{d\mu_i}
    = \begin{cases}
        \alpha_i, & \text{if } i = k, \\
        \alpha_i - \frac{1}{c(\sigma_i^2+\sigma_k^2)} \exp\left(\frac{-(\mu_i-\mu_k)^2}{2(\sigma_i^2+\sigma_k^2)}\right), & \text{if } i \ne k,
    \end{cases}
\end{equation*}
with $\alpha_i = \frac{2}{n} (\mu_i - y^{(i)})$ and $c = (n-1) \sqrt{2 \pi}$.\\

\begin{figure}[htp]
\centering
\includegraphics[width=\linewidth]{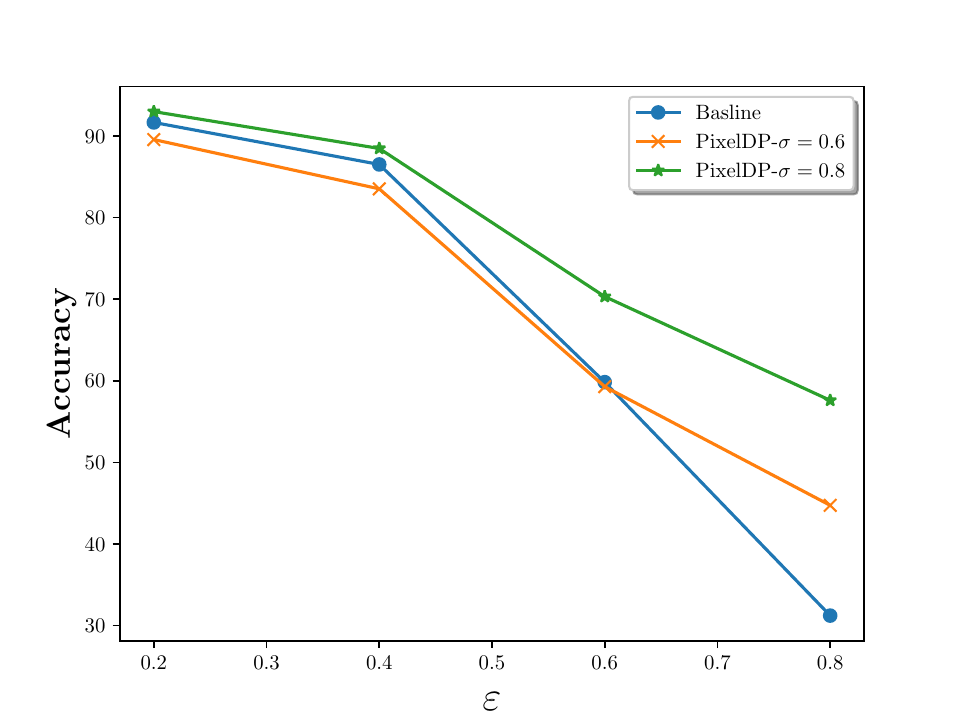}
\caption{Effect of the Adaptive attack on a inference-time randomised model for MNIST dataset }
\label{Fig22}
\end{figure}

This attack was carried out considering two different levels of noise, specifically $\sigma = 0.6$ and $\sigma = 0.8$ with one backward pass (FGSM method). Interestingly, the results depicted in Figure \ref{Fig22} revealed a vulnerability of the stochastic model to the adaptive attack. For $\sigma = 0.6$, the model is almost as vulnerable as a non-protected model.

\section{Related Work}

Several defense mechanisms were proposed to defend against adversarial attacks, we mainly distinguish:\\
\noindent
\textbf{Adversarial Training (AT).} 
AT is one of the most explored defenses against adversarial attacks. The main idea can be traced back to \cite{fgsm}, in which models were hardened by including adversarial examples in the training data set of the model. Nonetheless, AT is much more computationally intensive than training a model on the training data set only because generating evasive samples needs more computation.\\

\noindent
\textbf{Randomization-based Defenses.} These techniques use random noise at inference time to defend against adversarial attacks \cite{snP2019_certif,smooth,liu2017}. Liu et al. \cite{liu2017} suggest to randomize the entire DNN and predict using an ensemble of multiple copies of the DNN. Lecuyer et al. \cite{snP2019_certif} also suggest to add random noise to the first layer of the DNN and estimate the output by a Monte Carlo simulation. From a practical perspective, it is challenging for these works to scale and are limited with the necessity of MC simulation at inference time. 

\section{Discussion and concluding remarks}

In this paper, we propose a new approach to train adversarially robust models without the need of generating adversarial samples. Our proposition is based on a hybridation between adversarial training on the one hand, and randomization defenses on the other hand. In fact, while AT trains the model under an $\ell_p$-norm ball noise around the input samples to distance the decision boundary from the data distribution, we propose to optimize the model's parameters under stochastic behavior of the model itself to obtain the same objective.

To enable noise-aware training, we derived a closed form loss function that encapsulates the noise distribution propagated through the model. Additionally, we formulated a noise-aware gradient, which backpropagated to update the model's parameters. Once the model is trained, we tested the expectation model, i.e., without noise, at inference time.

We evaluate the model's accuracy under various adversarial attacks such as FGSM and PGD. Our experiments confirmed our initial intuition and showed that the proposed method trained robust models without adversarial examples, and without accuracy drop compared to baseline vanilla models. 

Interestingly, we also show that the proposed stochastic loss function can be used to generate efficient adversarial attacks against inference-time randomization based defenses.

One potential limitation of this approach is that it may require more computational resources than conventional training, as it involves optimizing an additional parameter and approximations. However, in contradiction to existing randmization techniques, the inference is deterministic and that the benefits in terms of robustness are significant. Overall, the results of our study suggest that incorporating the noise variance as a parameter in the neural network can be an effective defense mechanism against adversarial attacks.

Another finding we provide in the supplementary material suggests that the parameters of the injected noise within the model is also \textit{a learnable parameter} that can be integrated in the model training. In fact, instead of fixing $\sigma$, we consider it as a parameter of the model and we update it in the training process. Interestingly, the model did not converge to a deterministic model ($\sigma =0$), but rather to an "optimally stochastic" model. More details can be found in the supplementary materials.

Further research is needed to explore the full potential of this approach and its applicability to different types of neural networks and learning architectures.


{
    \small
    \bibliographystyle{ieeenat_fullname}
    \bibliography{main}
}
\clearpage
\setcounter{page}{1}
\maketitlesupplementary

\section{Can we optimize the noise as a parameter?}\label{sec:xyz}


The core paper is interested in stochastically training ML models under a fixed noise parameter (standard deviation), and inferring the trained model in a deterministic fashion (by taking the expectation, i.e.,  $\sigma_{inference}=0$). In this section, we explore the following question: \\

\noindent\textbf{\underline{Q} What would be the model's behavior if we consider $\sigma$ as a \textit{learnable parameter}? }

More specifically, if we train the model under noise, while updating the noise parameter during training the same as weights and biases, we want to investigate the correctness of the following hypothesis:

\begin{tcolorbox}[colback=blue!5!white,colframe=white]
\textit{\textbf{Hypothesis} -- $\mathcal{H}$: "If we train the model while optimizing $\sigma$, it converges to a deterministic model, i.e., finds that the minimization of the stochastic loss systematically converges to a $\sigma = 0$"}
\end{tcolorbox}
A way of conceptualizing this experiment is that we are expanding the dimensionality of the problem by introducing noise as a new dimension of the parameters' space.

To investigate $\mathcal{H}$, we train the Lenet-5 model under noise, while initializing $\sigma$ randomly (we did not witness any specific difference  made by the initialization). The update of $\sigma$ is simply made by chain rule to find $ \frac{\partial \operatorname{Loss}}{\partial \sigma}$. 
This will allow us to converge to the optimal values of parameters including noise standard deviation. If $\sigma$ converges to $0$ than $\mathcal{H}$ is verified. 

In a another subsequent setting, we update the loss function such that we minimize the stochastic loss under maximization of the noise itself. This setting is to explore the the maximum allowable noise while training the model.  In this scenario, the expression of the loss function will be as follows:
\begin{equation*}\label{eq:max_sigma}
    \operatorname{Loss} = \operatorname{MSE} - \sum_{\substack{i = 1 \\ i \neq k}}^n \frac{1}{n-1} \mathbb{P}\left(Y_{k} > Y_{i}\right)-\alpha \sigma^2,
\end{equation*}

While we use the previously expressed closed form of the stochastic loss and add $\sigma^2$ multiplied by an empirical regularization factor $\alpha$. In this analysis, we used $\alpha=0.25$ for the Bimodel.

The results are presented in Table \ref{t1}, where "Bimodel" denotes the model with the maximization objective and "Model" refers to the stochastic model without maximization of the noise, $\sigma_0$ is the initialisation and $\sigma^*$ is the value of noise standard deviation that the trained model converged to.  

Interestingly, even without maximisation of the noise, we noticed that the model converges to a non-zero $\sigma$ value, \textbf{which refutes the Hypothesis $\mathcal{H}$}.  

\begin{table}[hp]
    \centering
    \begin{tabular}{ |c|c|c|c| } 
    \hline
    Model & $\sigma_0$ & $\sigma^\star$ \\
    \hline
        Model   & 1.9 & 0.7  \\ 
        Bimodel & 1.9 & 0.79 \\ 
    \hline
    \end{tabular}
    \caption{Values of converged noise: BiModel denote the model with the maximization objective and Model denote the one without any maximization objective. }
    \label{t1}
\end{table}

\end{document}